\documentclass[10pt,journal,compsoc]{IEEEtran}
\ifCLASSOPTIONcompsoc
  \usepackage[nocompress]{cite}
\else
  \usepackage{cite}
\fi
\ifCLASSINFOpdf
\else
\fi

\usepackage{amsmath}
\usepackage{multirow} 
\usepackage{multicol}
\usepackage{arydshln}
\usepackage{subfigure}
\usepackage{graphicx}
\usepackage{url}
\usepackage[numbers]{natbib}
\usepackage{ragged2e}
\usepackage{subfigure}
\usepackage{pgfplots}
\usepackage{booktabs}
\usepackage{longtable}
\begin{document}
\title{SG-Net: Syntax Guided Transformer for Language Representation}

\author{Zhuosheng Zhang, Yuwei Wu, Junru Zhou, Sufeng Duan, Hai Zhao*, Rui Wang
		\IEEEcompsocitemizethanks{\IEEEcompsocthanksitem{This paper was partially supported by National Key Research and Development Program of China (No. 2017YFB0304100), Key Projects of National Natural Science Foundation of China (U1836222 and 61733011), Huawei-SJTU long term AI project, Cutting-edge Machine Reading Comprehension and Language Model (Corresponding author: Hai Zhao).}
		\IEEEcompsocthanksitem{Z. Zhang, Y. Wu, J. Zhou, S. Duan, H. Zhao, R. Wang are with the Department of Computer Science and Engineering, Shanghai Jiao Tong University, and also with Key Laboratory of Shanghai Education Commission for Intelligent Interaction and Cognitive Engineering, Shanghai Jiao Tong University, and also with MoE Key Lab of Artificial Intelligence, AI Institute, Shanghai Jiao Tong University. \protect\\ E-mail: \{zhangzs, will8821, zhoujunru, 1140339019dsf\}@sjtu.edu.cn, zhaohai@cs.sjtu.edu.cn, wangrui.nlp@gmail.com.}
    	\IEEEcompsocthanksitem{Part of this work was finished when Z. Zhang visited National Institute of Information and Communications Technology (NICT) and R. Wang was with NICT.}
		\IEEEcompsocthanksitem{Part of this study has been accepted as ``SG-Net: Syntax-Guided Machine Reading Comprehension'' \cite{Zhang2020sgnet} in the Thirty-Fourth AAAI Conference on Artificial Intelligence (AAAI 2020). This paper extends the previous syntax-guided attention method to natural language comprehension, inference, and generation tasks. We further conduct comprehensive experiments, to verify the effectiveness, as well as generalization ability on different benchmarks with thorough case studies.}}
	}

\IEEEtitleabstractindextext{%
\begin{abstract}
\justifying 
Understanding human language is one of the key themes of artificial intelligence. For language representation, the capacity of effectively modeling the linguistic knowledge from the detail-riddled and lengthy texts and getting rid of the noises is essential to improve its performance. Traditional attentive models attend to all words without explicit constraint, which results in inaccurate concentration on some dispensable words. In this work, we propose using syntax to guide the text modeling by incorporating explicit syntactic constraints into attention mechanisms for better linguistically motivated word representations. In detail, for self-attention network (SAN) sponsored Transformer-based encoder, we introduce syntactic dependency of interest (SDOI) design into the SAN to form an SDOI-SAN with syntax-guided self-attention. Syntax-guided network (SG-Net) is then composed of this extra SDOI-SAN and the SAN from the original Transformer encoder through a dual contextual architecture for better linguistics inspired representation. The proposed SG-Net is applied to typical Transformer encoders. Extensive experiments on popular benchmark tasks, including machine reading comprehension, natural language inference, and neural machine translation show the effectiveness of the proposed SG-Net design.
\end{abstract}

}

\maketitle

\IEEEdisplaynontitleabstractindextext

\IEEEpeerreviewmaketitle

\IEEEraisesectionheading{\section{Introduction}\label{sec:introduction}}
\IEEEPARstart{T}{eaching} machines to read and comprehend human languages is a long-standing goal of artificial intelligence, where the fundamental is language representation. Recently, much progress has been made in general-purpose language representation that can be used across a wide range of tasks \cite{radford2018improving,devlin2018bert,zhang2019semantics,zhou2019limit,zhang2020neural}. Understanding the meaning of texts is a prerequisite to solve many natural language understanding (NLU) problems, such as machine reading comprehension (MRC) based question answering \cite{Rajpurkar2018Know}, natural language inference (NLI) \cite{Bowman2015A}, and neural machine translation (NMT) \cite{bahdanau2014neural}, all of which require a rich and accurate representation of the meaning of a sentence.

For language representation, the first step is to encode the raw texts in vector space using an encoder. The dominant language encoder has evolved from recurrent neural networks (RNN) to Transformer \cite{vaswani2017attention} architectures. With stronger feature extraction ability than RNNs, Transformer has been widely used for encoding deep contextualized representations \cite{Peters2018ELMO,radford2018improving,devlin2018bert}. After encoding the texts, the next important target is to model and capture the text meanings.
Although a variety of attentive models \cite{bahdanau2014neural,wang2017learning} have been proposed to implicitly pay attention to key parts of texts, most of them, especially global attention methods \cite{bahdanau2014neural} equally tackle each word and attend to all words in a sentence without explicit pruning and prior focus, which would result in inaccurate concentration on some dispensable words \cite{Mudrakarta2018Did}.

\begin{figure}
	\centering
	\includegraphics[width=0.49\textwidth]{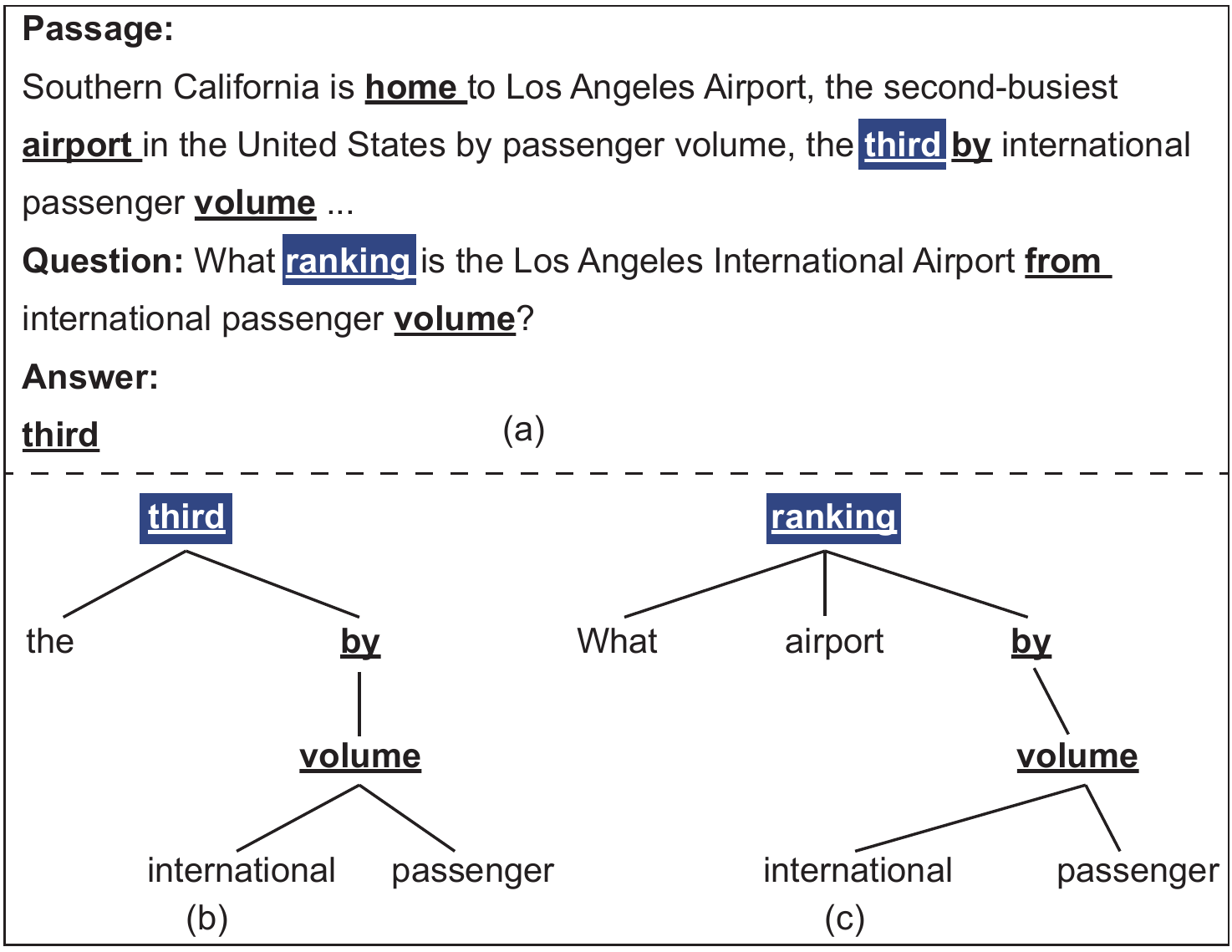}
	\caption{(a) Example of syntax-guided span-based QA. The SDOI of each word consists of all its ancestor words. (b-c) The dependency parsing tree of the given passage sentence and question.}
	\label{exp_dep}
\end{figure}

Taking machine reading comprehension based question answering (QA) task \cite{Rajpurkar2016SQuAD,Rajpurkar2018Know} (Figure \ref{exp_dep}) as an example, we observe that the accuracy of MRC models decreases when answering long questions (shown in Section \ref{sec:long}). Generally, if the text is particularly lengthy and detailed-riddled, it would be quite difficult for a deep learning model to understand as it suffers from noise and pays vague attention to the text components, let alone accurately answering questions \cite{zhang2018OneShot,zhang2018SubMRC,zhang2018modeling}. In contrast, existing studies have shown that machines could read sentences efficiently by taking a sequence of fixation and saccades after a quick first glance \cite{P17-1172} or referring to compressed texts \cite{li2020explicit}, which inspire us to integrate structured information into a language representation model to obtain a hierarchical and focused representation. The most common source for structure information is syntactic parsing \cite{shen2018neural}.

Incorporating human knowledge into neural models is one of the major research interests of artificial intelligence. Recent Transformer-based deep contextual language representation models have been widely used for learning universal language representations from large amounts of unlabeled data, achieving dominant results in a series of NLU benchmarks \cite{Peters2018ELMO,radford2018improving,devlin2018bert,yang2019xlnet,liu2019roberta,lan2019albert}. However, they only learn from plain context-sensitive features such as character or word embeddings, with little consideration of explicitly extracting the hierarchical dependency structures that are entailed in human languages, which can provide rich dependency hints for language representation. 

Besides, as a common phenomenon in language representation, an input sequence always consists of multiple sentences. Nearly all of the current attentive methods and language representation models regard the input sequence as a whole, e.g., a passage, with no consideration of the inner linguistic structure inside each sentence. This would result in process bias caused by much noise and a lack of associated spans for each concerned word. 

All these factors motivate us to seek for an informative method that can selectively pick out important words by only considering the related subset of words of syntactic importance inside each input sentence explicitly. With the guidance of syntactic structure clues, the syntax-guided method could give more accurate attentive signals and reduce the impact of the noise brought about by lengthy sentences. 

So far, we have two types of broadly adopted contextualized encoders for building sentence-level representation, i.e., RNN-based and Transformer-based \cite{vaswani2017attention}. The latter has shown its superiority, which is empowered by a self-attention network (SAN) design. In this paper, we extend the self-attention mechanism with syntax-guided constraint, to capture syntax related parts with each concerned word. Specifically, we adopt pre-trained dependency syntactic parse tree structure to produce the related nodes for each word in a sentence, namely syntactic dependency of interest (SDOI), by regarding each word as a child node and the SDOI consists all its ancestor nodes and itself in the dependency parsing tree. An example is shown in Figure \ref{exp_dep}. 

To effectively accommodate such SDOI information, we propose a novel syntax-guided network (SG-Net), which fuses the original SAN and SDOI-SAN, to provide more linguistically inspired representation to comprehend language hierarchically. Our method allows the Transformer to learn more interpretable attention weights that better explain how the model processes and comprehends the natural language. 

The contribution of this paper is three-fold:

1) We propose a novel syntax-guided network (SG-Net) that induces explicit syntactic constraints to Transformer architectures for learning structured representation. To our best knowledge, we are the first to integrate syntactic relationships  as attentive guidance for enhancing state-of-the-art SAN in the Transformer encoder.

2) Our model boosts the baseline Transformer significantly on three typical advanced natural language understanding tasks. The attention learning shows better interpretability since the attention weights are constrained to follow the induced syntactic structures.  By visualizing the self-attention matrices, our model provides  the  information  that  better  matches the human intuition about hierarchical structures than the original Transformer.

3) Our proposed method is lightweight and easy to implement. We can easily apply syntax-guided constraints in an encoder by adding an extra layer with the SDOI mask, which makes it easy enough to be applicable to other systems.

\section{Background}
\subsection{Language Representation}
Language representation is the foundation of deep learning methods for natural language processing and understanding. As the basic unit of language representation, learning word representations has been an active research area, and aroused great research interests for decades, including non-neural \cite{brown1992class,ando2005framework,blitzer2006domain} and neural methods \cite{mikolov2013distributed,pennington2014glove}. 

Distributed representations have been widely used as a standard part of natural language processing (NLP) models due to the ability to capture the local co-occurrence of words from large scale unlabeled text \cite{mikolov2013distributed}. However, when using the learned embedding for NLP tasks, these approaches for word vectors only involve a single, context independent representation for each word with a word-vectors lookup table, with litter consideration of contextual encoding in sentence level. Thus recently introduced contextual language representation models including ELMo \cite{Peters2018ELMO}, GPT \cite{radford2018improving} and BERT \cite{devlin2018bert} fill the gap by strengthening the contextual sentence modeling for better representation,\footnote{Strictly speaking, the pre-trained models are not traditionally-defined language models when in use, as the former are mainly used as encoders to vectorize texts, which we call \textit{language (representation) model} following the formulation in previous works \cite{devlin2018bert}.} showing powerful to boost NLU tasks to reach new high performance. Two stages of training are adopted in these models: firstly, pre-train a model using language model objectives on a large-scale text corpus, and then fine-tune the model (as an pre-trained encoder with simple linear layers) in specific downstream NLP tasks (e.g., text classification, question answering, natural language inference). Among these contextualized language representation  models, BERT uses a different pre-training objective, masked language model, which allows capturing both sides of context, left and right. Besides, BERT also introduces a \emph{next sentence prediction} task that jointly pre-trains text-pair representations. The latest evaluation shows that BERT is powerful and convenient for downstream NLU tasks.

\begin{figure*}[htb]
	\centering
	\includegraphics[width=0.9\textwidth]{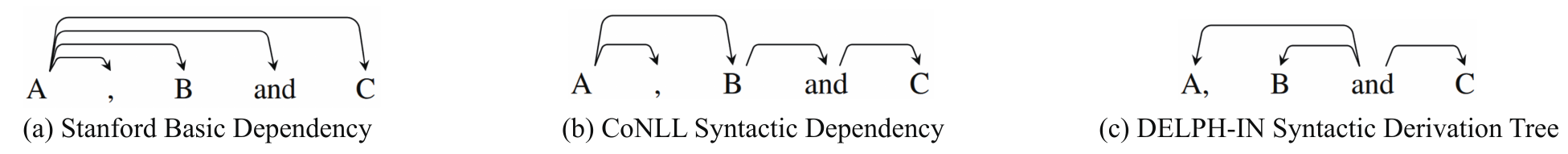}
	\caption{\label{fig:dp_types} Example of dependency formats.}
\end{figure*}

The major technical improvement over traditional embeddings of these newly proposed language representation models is that they focus on extracting context-sensitive features from language models. When integrating these contextual word embeddings with existing task-specific architectures, ELMo helps boost several major NLP benchmarks \cite{Peters2018ELMO} including question answering on SQuAD, sentiment analysis \cite{socher2013recursive}, and named entity recognition \cite{sang2003introduction}, while BERT especially shows effective on language understanding tasks on GLUE \cite{wang2018glue}, MultiNLI \cite{nangia2017repeval} and SQuAD \cite{devlin2018bert}. In this work, we follow this line of extracting context-sensitive features and take pre-trained BERT as our backbone encoder for jointly learning explicit context semantics.

However, the pre-trained language representation models only exploit plain context-sensitive features such as character or word embeddings. They rarely consider incorporating explicit, structured syntactic information, which can provide rich dependency information for language representation. To promote the ability to model  the structured dependency of words in a sentence, we are inspired to incorporate extra syntactic constraints into the multi-head attention mechanism in the dominant language representation models.
The latest evaluation shows that BERT is powerful and convenient for downstream tasks. Following this line, we extract context-sensitive syntactic features and take pre-trained BERT as our backbone encoder to verify the effectiveness of our proposed SG-Net.

\subsection{Syntactic Structures}
Syntactic structure consists of lexical items, linked by binary asymmetric relations called dependencies \cite{nivre2005inductive}. For parsing dependency structures, the salient resources are Treebanks, which are collections of sentences that have been manually annotated with a correct syntactic analysis and part-of-speech (PoS) tags. Among them, the Penn Treebank (PTB) is an annotated corpus consisting of 50$k$ sentences with over 4.5 million words of American English \cite{marcus1993building}. 

Generally, three annotation formats are used for dependency parsing \cite{ivanova2013survey}, including two classic  representations  for dependency parsing, namely, \textit{Stanford Basic} (SB) and \textit{CoNLL Syntactic Dependencies} (CD), and bilexical dependencies  from  the \textit{head-driven phrase structure grammar} (HPSG) English Resource Grammar (ERG), so-called \textit{DELPH-IN Syntactic Derivation Tree} (DT). The annotation comparison is shown in Figure \ref{fig:dp_types}. HPSG is a highly lexicalized, constraint-based grammar developed by 
\citeauthor{pollard1994head}~(\citeyear{pollard1994head}) \cite{pollard1994head}, which enjoys a uniform formalism representing rich contextual syntactic and  even semantic meanings. HPSG divides language symbols into categories of different types, such as vocabulary, phrases, etc. The complete language symbol which is a  complex  type  feature  structure  represented  by attribute value matrices (AVMs) includes phonological, syntactic, and semantic properties, the valence  of  the  word  and  interrelationship  between various components of the phrase structure. Based on the above advantages of modeling word interrelationships of HPSG, we use the HPSG format to obtain the syntactic dependencies in this work.

There are three major classes of parsing models \cite{kubler2009dependency}, \textit{transition-based} \cite{dyer2015transition,zhang2011transition}, \textit{graph-based} \cite{wang2016graph,zhang2016probabilistic}, and \textit{grammar-based models} \cite{blunsom2010unsupervised,metheniti2019identifying}. 
Recently, dependency syntactic parsing has been further developed with neural network and attained new state-of-the-art results \cite{zhang-etal-2016-probabilistic,li-etal-2018-seq2seq,Ma2018Stack,li2020global}.
Benefiting from the highly accurate parser, neural network models could enjoy even higher accuracy gains by leveraging syntactic information rather than ignoring it \cite{chen2017neural2,chen2017neural,chen2018syntax,wang2019tree,duan2019ialp}. 

Syntactic dependency parse tree provides a form that is capable of indicating the existence and type of linguistic dependency relation among words, which has been shown generally beneficial in various natural language understanding tasks \cite{bowman2016fast}. To effectively exploit syntactic clue, our early work \cite{chen2018syntax} extended local attention with syntax-distance constraint in RNN encoder by focusing on syntactically related words with the predicted target words for neural machine translation; \citeauthor{kasai2019syntax}~(\citeyear{kasai2019syntax}) \cite{kasai2019syntax} absorbed parse tree information by transforming dependency labels into vectors and simply concatenated the label embedding with word representation. However, such simplified and straightforward processing would result in higher dimensions of the joint word and label embeddings and is too coarse to capture contextual interactions between the associated labels and the mutual connections between labels and words. This inspires us to seek for an attentive way to enrich the contextual representation from the syntactic source. A related work is from \citeauthor{strubell2018linguistically}~(\citeyear{strubell2018linguistically}) \cite{strubell2018linguistically}, which proposed to incorporate syntax with multi-task learning for semantic role labeling. However, their syntax is incorporated by training one extra attention head to attend to syntactic ancestors for each token while we use all the existing heads rather than adding an extra one. Besides, this work is based on the remarkable representation capacity of recent language representation models such as BERT, which have been suggested to be endowed with some syntax to an extent \cite{clark2019does}. Therefore, we are motivated to apply syntactic constraints through syntax guided method to prune the self-attention instead of purely adding dependency features.

In this work, we form a general approach to benefit from syntax-guided representations, which is the first attempt for the SAN architecture improvement in Transformer encoder to our best knowledge. The idea of updating the representation of a word with information from its neighbors in the dependency tree, which benefits from explicit syntactic constraints, is well linguistically motivated.

Our work differentiates from previous studies by both sides of technique and motivation: 

1) Existing work \cite{chen2018syntax,wu2018phrase} only aimed to improve the dependency modeling in RNN while this is the pioneering work to integrate the parsing structure into pre-trained Transformer with a new methodology as SDOI mask. Our practice is much more difficult as Transformer is a stronger baseline with some ability to capture dependency itself;

2) Previous work majorly focused on similar fundamental linguistic tasks \cite{strubell2018linguistically}. In contrast, we are motivated to apply linguistics (e.g., syntax) to more advanced NLU tasks (e.g., MRC), which are more complex and closer to AI. For NLU, pre-trained Transformers-based language representation models (e.g., BERT) have been widely used. It would be more beneficial for downstream tasks by taking them as the backbone, instead of focusing on RNNs that are less important for NLU recently. The advance would be potential to inspire the applications in other NLP and NLU tasks.

For the application, our method is lightweight and easy to cooperate with other neural models. Without the need for architecture modifications in existing models, we can easily apply syntax-guided constraints in an encoder by adding an extra layer fed by the output representation of the encoder and the SDOI mask, which makes it easy enough to be applicable to other systems.
\begin{figure*}
	\centering
	\includegraphics[width=1.0\textwidth]{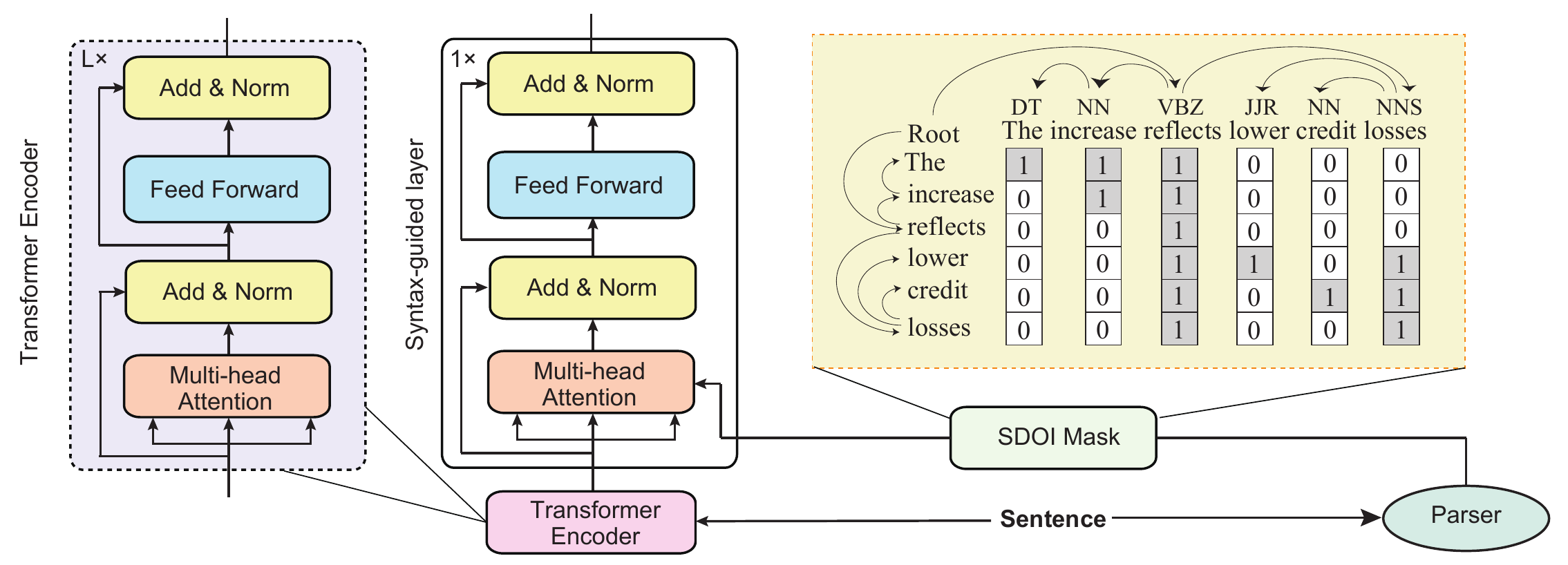}
	\caption{\label{fig:overview} Overview of the syntax-guided network.}
\end{figure*}
\section{Syntax-Guided Network}
Our goal is to design an effective neural network model that makes use of structured linguistic information as effectively as possible. We first present the general syntax-guided attentive architecture, building upon the recent advanced Transformer-based encoder as task-agnostic architecture and then fit with downstream task-specific layers for downstream language comprehension tasks.\footnote{Note that our method is not limited to cooperate with BERT in our actual use, but any encoder with a self-attention network (SAN) architecture.}

Figure \ref{fig:overview} depicts the whole architecture of our model. Our model first directly takes the output representations from an SAN-empowered Transformer-based encoder, then builds a syntax-guided SAN from the SAN representations. At last, the syntax-enhanced representations are fused from the syntax-guided SAN and the original SAN and passed to task-specific layers for final predictions. Table  \ref{tab:symbol}  summarizes  the  notations and symbols used in this article.

\begin{table}
\centering
\caption{\label{tab:symbol} Our notations and symbols.}
	{
		\begin{tabular}{l l}
			\hline
			\hline
			Symbol & Meaning\\
			\hline
            T & Input text sequence \\
            X & Embedding of the input sequence \\
            \textbf{H} & Contextual representation from the encoder \\
            $\mathcal{M}$ & SDOI mask \\
            H' & Output of the syntax-guided layer \\
            $\bar H$ & Output of the context aggregation \\
            FFN & Feedforward Network \\
            MultiHead & Multihead attention layer \\
			\hline
			\hline
		\end{tabular}
	}
	
\end{table}

\subsection{Transformer Encoder}
The raw text sequences are firstly represented as embedding vectors to feed an encoder (e.g., a pre-trained language representation model such as BERT \cite{devlin2018bert}).
	The input sentence is first tokenized to word pieces (subword tokens). Let $T=\{t_1,\dots,t_n\}$ denote a sequence of subword tokens of length $n$. 
	For each token, the input embedding is the sum of its token embedding, position embedding, and token-type embedding.\footnote{For BERT-like models \cite{devlin2018bert,lan2019albert}, they often take sequence pairs (e.g., passage + question) which are packed into a single sequence as input. Token type here is used to indicate whether each token belongs to the first sequence, or the second one.}
	Let $X = \{x_1, \dots, x_n\}$ be the embedding of the sequence, which are features of encoding sentence words of length $n$. The input embeddings are then fed into the multi-head attention layer to obtain the contextual representations.
	
	The embedding sequence $X$ is processed to a multi-layer bidirectional Transformer for learning contextual representations, which is defined as 
	\begin{align}\label{eq:mutihead}
	\textbf{H} = \textup{FFN}(\textup{MultiHead}(K,Q,V)),
	\end{align}
	where K,Q,V are packed from the input sequence representation $X$. As the common practice, we set $K=Q=V$ in the implementation.
	
	In detail, the function of Eq. (\ref{eq:mutihead}) is defined as follows.
	Let $X^{l} = \{x^{l}_{1}, \dots, x^{l}_{n} \}$ be the features of the $l$-th layer. In the $(l+1)$-th layer, the corresponding features $x^{l+1}$ are computed by
	\begin{align}
	\tilde{h}_{i}^{l+1} &= \sum_{m=1}^{M}W_{m}^{l+1}\left \{ \sum_{j=1}^{N}A_{i,j}^{m}\cdot V_{m}^{l+1}x_{j}^{l} \right \},\label{mha} \\
	h_{i}^{l+1} &= \textup{LayerNorm}(x_{i}^{l}+\tilde{h}_{i}^{l+1}), \\
	\tilde{x}_{i}^{l+1} &= W_{2}^{l+1}\cdot \textup{GELU}(W_{1}^{l+1}h_{i}^{l+1}+ b_{1}^{l+1}) + b_{2}^{l+1}, \label{ffn}\\
	x_{i}^{l+1} &= \textup{LayerNorm}(h_{i}^{l+1} + \tilde{x}_{i}^{l+1}),
	\end{align}
	where $m$ 
	is the index of the attention heads, and $A_{i,j}^{m}\propto \mathrm{exp}[(Q_{m}^{l+1}x_{i}^{l})^{\top }(K_{m}^{l+1}x_{j}^{l})]$ denotes the attention weights between elements $i$ and $j$ in the $m$-th head, which is normalized by $\sum_{j=1}^{N}A_{i,j}^{m}=1$. $W_{m}^{l+1}, Q_{m}^{l+1}, K_{m}^{l+1}$ and $V_{m}^{l+1}$ are learnable weights for the $m$-th attention head, $W_{1}^{l+1}, W_{2}^{l+1}$ and $b_{1}^{l+1},b_{2}^{l+1}$ 
	are learnable weights and biases, respectively. 
	
	For the following part, we use $\textbf{H} = \{h_1, \dots, h_n\}$ to denote the last-layer hidden states of the input sequence.

\subsection{Syntax-Guided self-attention Layer}  \label{SG_ATT}

Our syntax-guided representation is obtained through two steps. Firstly, we pass the encoded representation from the Transformer encoder to a syntax-guided self-attention layer. Secondly, the corresponding output is aggregated with the original encoder output to form a syntax-enhanced representation. It is designed to incorporate the syntactic tree structure information inside a multi-head attention mechanism to indicate the token relationships of each sentence, which will be demonstrated as follows.

In this work, we first pre-train a syntactic dependency parser to annotate the dependency structures for every sentence, which are then fed to SG-Net as guidance of token-aware attention. Details of the pre-training process of the parser are reported in Section \ref{imp}. 


To use the relationship between head word and dependent words provided by the syntactic dependency tree of sentence, we restrain the scope of attention only between word and all of its ancestor head words.\footnote{We extend the idea of using parent in \citeauthor{strubell2018linguistically}~(\citeyear{strubell2018linguistically}) \cite{strubell2018linguistically} to ancestor for a wider receptive range.} In other words, we would like to have each word only attend to words of syntactic importance in a sentence, the ancestor head words in the view of the child word. As the SDOI mask shown in Figure \ref{fig:overview}, instead of taking attention with each word in whole passage, the word \textit{credit} only makes attention with its ancestor head words \textit{reflects} and \textit{losses} and itself in this sentence, which means that the SDOI of \textit{credit} contains \textit{reflects}, \textit{losses} along with itself. Language representation models usually use special tokens, such as [CLS], [SEP], and [PAD] used in BERT, and $<$PAD$>$, $<$/s$>$ tokens used in Transformer for NMT. The SDOI of these special tokens is themselves alone in our implementation, which means these tokens will only attend to themselves in the syntax-guided self-attention layer.

Specifically, given input token sequence $S=\{s_1,s_2,...,s_n\}$ where $n$ denotes the sequence length, we first use syntactic parser to generate a dependency tree. Then, we derive the ancestor node set $P_i$ for each word $s_i$ according to the dependency tree.  Finally, we learn a sequence of SDOI
mask $\mathcal{M}$, organized as $n*n$ matrix, and elements in each row denote the dependency mask of all words to the row-index word.
\begin{equation}
\mathcal{M}[i,j] = \left\{\begin{matrix}
1, & {\text{ if } j} \in {P_i}\text{ or }j=i \\ 
0, & \text{otherwise}.
\end{matrix}\right.
\end{equation}%

Obviously, if $\mathcal{M}[i,j]=1$, it means that token $s_i$ is the ancestor node of token $s_j$. As the example shown in Figure \ref{fig:overview}, the ancestors of \textit{credit} ($i$=4) are \textit{reflects} ($j$=2), \textit{losses} ($j$=5) along with itself ($j$=4); therefore, $\mathcal{M}[4, (2,4,5)]=1$ and $\mathcal{M}[4, (0,1,3)]=0$.

We then project the last layer output $H$ from the vanilla Transformer into the distinct key, value, and query representations of dimensions $L\times d_k$, $L\times d_q$, and $L\times d_v$, respectively, denoted $K'_i, Q'_i$ and $V'_i$ for each head $i$.\footnote{In SG-Net, there is only one extra attention layer compared with multi-layer Transformer architecture. The total numbers of our model and the baseline Transformer are very close, e.g., taking BERT as the baseline backbone, the parameters are 347M (+SG layer) vs 335M. } Then we perform a dot product to score key-query pairs with the dependency of interest mask to obtain attention weights of dimension $L\times L$, denoted $A_i'$:
\begin{equation}
{A_i}' = \textup{Softmax} \left( {\frac{{\mathcal{M}\cdot \left( {{Q_i}'{K_i}'^T} \right)}}{{\sqrt {{d_k}} }}} \right).
\end{equation}

We then multiply attention weight $A_i'$ by $V_i'$ to obtain the syntax-guided token representations:
\begin{equation}
W_i' = A_i'V_i'.	
\end{equation}

Then $W_i'$ for all heads are concatenated and passed through a feed-forward layer followed by GeLU activations \cite{hendrycks2016gaussian}. After passing through another feed-forward layer, we apply a layer normalization to the sum of output and initial representation to obtain the final representation, denoted as $H'=\{h'_1,h'_2,...,h'_n\}$.


\subsection{Dual Context Aggregation}
 Considering that we have two representations now, one is $H =\{h_1,h_2,...,h_n\}$  from the Transformer encoder, the other is $H'=\{h'_1,h'_2,...,h'_n\}$ from syntax-guided layer from the above part. Formally, the final model output of our SG-Net $\bar H=\{\bar h_1,\bar h_2,...,\bar h_n\}$ is computed by:
\begin{equation}
\bar h_i = \alpha h_i + (1-\alpha)h'_i. \\
\end{equation}

\subsection{Task-specific Adaptation} \label{mrc_models}
\subsubsection{Machine Reading Comprehension}
We focus on two types of reading comprehension tasks, i.e., \emph{span-based} and \emph{multi-choice} style which can be described as a tuple $<P, Q, A>$ or $<P, Q, C, A>$ respectively, where $P$ is a passage (context), and $Q$ is a query over the contents of $P$, in which a span or choice $C$ is the right answer $A$. For the span-based one, we implemented our model on SQuAD 2.0 task that contains unanswerable questions. Our system is supposed to not only predict the start and end position in the passage $P$ and extract span as answer $A$ but also return a null string when the question is unanswerable. For the multi-choice style, the model is implemented on the RACE dataset which is requested to choose the right answer from a set of candidate ones according to given passage and question.

Here, we formulate our model for both of the two tasks and feed the output from the syntax-guided network to task layers according to the specific task. Given the passage $P$, the question $Q$, and the choice $C$ specially for RACE, we organize the input $X$ for the encoder as the following two sequences.

\begin{figure}[htb]
	\centering
	\includegraphics[width=0.42\textwidth]{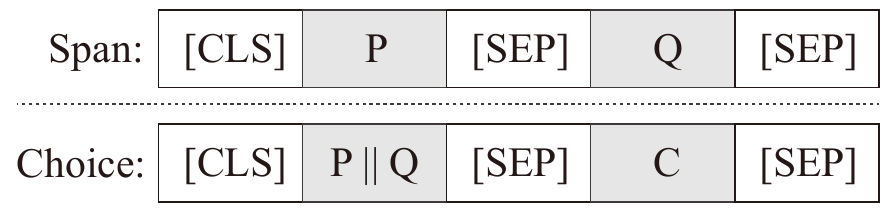}
\end{figure}
\noindent where $||$ denotes concatenation. 

In this work, pre-trained BERT is adopted as our detailed implementation of the Transformer encoder. Thus the sequence is fed to the BERT encoder mentioned above to obtain the contextualized representation $\textit{H}$, which is then passed to our proposed syntax-guided self-attention layer and aggregation layer to obtain the final syntax-enhanced representation $\bar{H}$. To keep simplicity, the downstream task-specific layer basically follows the implementation of BERT. We outline below to keep the integrity of our model architecture. For the span-based task, we feed $\bar{H}$ to a linear layer and obtain the probability distributions over the start and end positions through a softmax. For the multi-choice task, we feed it into the classifier to predict the choice label for the multi-choice model.

$\bullet$\textbf{SQuAD 2.0}
For SQuAD 2.0, our aim is a span of answer text; thus we employ a linear layer with SoftMax operation and feed $\bar H$ as the input to obtain the start and end probabilities, $s$ and $e$:
\begin{equation}
s, e = \textup{SoftMax}(\textup{Linear}(\bar H)).
\end{equation}


The training objective of our SQuAD model is defined as cross entropy loss for the start and end predictions, 
\begin{equation}
\begin{split}
\mathcal{L}_{has} = y_s \log{s} + y_e \log{e}.\\
\end{split}
\end{equation}
For prediction, given output start and end probabilities $s$ and $e$, we calculate the has-answer score $score_{has}$ and the no-answer score $score_{null}$:
\begin{equation}
\begin{split}
score_{has} & =\max (s_k + e_l),1 < k \le l \le n, \\
score_{null} & = s_1+e_1,
\end{split}
\end{equation}

We obtain a difference score $score_{diff}$ between $score_{has}$ and $score_{null}$ as the final score. A threshold $\delta$ is set to determine whether the question is answerable, which is heuristically computed with dynamic programming according to the development set. The model predicts the answer span that gives the has-answer score if the final score is above the threshold, and null string otherwise.

We find that adding an extra answer verifier module could yield a better result, which is trained in parallel only to determine whether question is answerable or not. The logits of the verifier are weighted with $score_{null}$ to give the final predictions. In detail, we employ a parallel MRC model (i.e., SG-Net) whose training objective is to measure the answerablity of the given questions. The verifier's function can be implemented as a cross-entropy loss. The pooled first token (the special symbol, \texttt{[CLS]}) representation $\bar h_{1} \in \bar H$, as the overall representation of the sequence, is passed to a fully connection layer to get classification logits $\hat{y}_{i}$ composed of answerable ($logit_{ans}$) and unanswerable ($logit_{na}$) elements. We use cross entropy as training objective: 
\begin{equation}
\mathcal{L}_{ans} = -\frac{1}{N}\sum_{i=1}^{N}\left [y_{i}\log\hat{y}_{i} + (1-y_{i})\log(1-\hat{y}_{i} )\right ],
\end{equation}
where $\hat{y}_{i} \propto \textup{SoftMax}(\textup{Linear}(\bar h_{1}))$ denotes the prediction and $y_{i}$ is the target indicating whether the question is answerable or not. $N$ is the number of examples. We calculate the difference as the new verification score: $score_{ext} = logit_{na} - logit_{ans}$. The no-answer score is calculated as:
\begin{equation}
score_{final} = \beta_{1} score_{diff} + \beta_{2} score_{ext},
\end{equation}
where $\beta_{1}$ and $\beta_{2}$ are weights. Our model predicts the answer span if $score_{final} > \delta$, and null string otherwise.

$\bullet$\textbf{RACE}
As discussed in \citeauthor{devlin2018bert}~(\citeyear{devlin2018bert}) \cite{devlin2018bert}, the pooled representation explicitly includes classification information during the pre-training stage of BERT. We expect the pooled to be an overall representation of the input. Thus, the first token representation $\bar h_0$ in $\bar H$ is picked out and is passed to a feed-forward layer to give the prediction $p$. For each instance with $n$ choice candidates, we update model parameters according to cross-entropy loss during training and choose the one with the highest probability as the prediction when testing. 
The training objectives of our RACE model is defined as, 
\begin{equation}
L(\theta) = -\frac{1}{N}\sum_i y_i \log{p}_i,
\end{equation}
where $p_{i}$ denotes the prediction, $y_{i}$ is the target, and $i$ denotes the data index.

\subsubsection{Natural Language Inference}
Since the natural language inference task is modeled as $n$-label classification problem, the model implementation is similar to RACE. The first token representation $\bar h_0$ in $\bar H$ is picked out and is passed to a feed-forward layer to give the prediction $p$. For each example with $n$ labels (e.g., \textit{entailment}, \textit{neutral} and \textit{contradiction}), we update model parameters according to cross-entropy loss during training and choose the one with highest probability as the prediction when testing.

\subsubsection{Neural Machine Translation}
Intuitively, NMT aims to produce a target word sequence with the same meaning as the source sentence, which is a natural language generation task. A Transformer NMT model consists of an encoder and a decoder, which fully rely on self-attention networks (SANs), to translate a sentence in one language into another language with equivalent meaning.

We introduce the syntax-guided self-attention layer in the encoder in the same way as the other tasks. The SAN of decoder then uses both $\bar H_i$ and target context hidden state $H_{tgt}$ to learn the context vector $o_i$ by ``encoder-decoder attention":
\begin{equation}
\begin{split}
c_{i} & =\textup{FFN}(\textup{MultiHead}(H_{tgt},H_{tgt},\bar H_i)),\\
o_i &= c_i + H_{tgt}.
\end{split}
\end{equation}

Finally, the context vector $o_i$ is used to compute translation probabilities of the next target word $y_i$ by  a linear, potentially multi-layered function:
\begin{equation}
P(y_i\mid y_{<i},x)\propto \textup{SoftMax}({L}_{0}\textup{GELU}({L}_{w}o_{i})),
\end{equation}
where $L_{0}$ and $L_{w}$ are projection matrices.

\section{Task Setup}
\subsection{Reading Comprehension}
Our experiments and analysis are carried on two data sets, involving span-based and multi-choice MRC and we use the fine-tuned cased BERT (whole word masking) as the baseline.

$\bullet$ \textbf{Span-based MRC}. As a widely used MRC benchmark dataset, SQuAD 2.0  \cite{Rajpurkar2018Know} combines the 100,000 questions in SQuAD 1.1 \cite{Rajpurkar2016SQuAD} with over 50,000 new, unanswerable questions that are written adversarially by crowdworkers to look similar to answerable ones. For the SQuAD 2.0 challenge, systems must not only answer questions when possible, but also abstain from answering when no answer is supported by the paragraph. Two official metrics are selected to evaluate the model performance: Exact Match (EM) and a softer metric F1 score, which measures the weighted average of the precision and recall rate at a character level.

$\bullet$\textbf{Multi-choice MRC}.
Our multi-choice MRC is evaluated on Large-scale ReAding Comprehension Dataset From Examinations (RACE) dataset \cite{lai2017race}, which consists of two subsets: RACE-M and RACE-H corresponding to middle school and high school difficulty levels. RACE contains 27,933 passages and 97,687 questions in total, which is recognized as one of the largest and most difficult datasets in multi-choice MRC. The official evaluation metric is accuracy.

\subsection{Natural Language Inference}

Natural language inference (NLI) is proposed to serve as a benchmark for natural language understanding and inference, which is also known as recognizing textual entailment (RTE). In this task, a model is presented with a pair of sentences and asked to judge the relationship between their meanings, including entailment, neutral, and contradiction. \citeauthor{Bowman2015A}~(\citeyear{Bowman2015A}) \cite{Bowman2015A} released Stanford Natural language Inference (SNLI) dataset, which is a collection of 570$k$ human-written English sentence pairs manually labeled for balanced classification with the labels \textit{entailment}, \textit{contradiction}, and \textit{neutral}. As a high-quality and large-scale benchmark, the SNLI corpus has inspired various significant work \cite{mou2015natural,chen2017recurrent,ghaeini2018dr,liu2019multi,zhang2019explicit,zhang2019semantics}. 

\subsection{Neural Machine Translation}
The proposed  NMT  model  was  evaluated  on  the  WMT14 English-to-German (EN-DE) translation task, which is a  standard  large-scale  corpus  for NMT  evaluation. For the translation task, 4.43M bilingual sentence pairs of the WMT14 dataset were used as training data, including Common Crawl, News Commentary, and Europarl v7.  The \textit{newstest2013} and \textit{newstest2014} datasets were used as the dev set and test set, respectively. The evaluation metric is BLEU score.

\section{Implementation} 
\label{imp}

\subsection{Syntactic Parser}
For the syntactic parser, we adopt the dependency parser from \citeauthor{zhou2019head}~(\citeyear{zhou2019head}) \cite{zhou2019head} by joint learning of  constituent parsing \cite{Kitaev-2018-SelfAttentive} using BERT as sole input which achieves very high accuracy: 97.00\% UAS and 95.43\% LAS on the English dataset Penn Treebank (PTB) \cite{MarcusJ93-2004} test set.\footnote{We report the results without punctuation of the labeled and unlabeled attachment scores (LAS, UAS).} 
Due to the fact that the parsing corpus is annotated in word-level, the parser can only produce the word-level tree structures. Therefore, previous works that employ syntactic information conduct experiments in word-level \cite{chen2018syntax,wang2018syntax}. However, the word-level setting will harm the baseline performance as the recent dominant models commonly benefit from subword tokens as input due to the efficiency and effectiveness as the minimal language unit \cite{Sennrich2015Neural,Bojanowski2016Enriching,zhang2019effective}. 

In SG-Net, to avoid the harm to the advanced subword-based strong Transformer baseline, we transform the word-level parsing structures into subword-level to ensure that the SDOI mask is in the same shape as the subword sequence shape. The transformation is based on the criterion that if the word is segmented into subwords, the subwords share the same tag as the original word. Alternatively, we tried to use the other method: If a word is segmented into $n$ subwords, the [1:n] subwords are regarded as the children of the first subword. We find that taking both of the criteria yields very similar final results, thus only report the former one.

Note the above parsing work for sentences is done in data preprocessing, and our parser is not updated with the following downstream models. For passages in the MRC tasks, the paragraphs are parsed sentence by sentence.

\subsection{Downstream-task Models}

For MRC model implementation, we adopt the Whole Word Masking BERT and ALBERT as the baselines.\footnote{\url{https://github.com/huggingface/transformers}. Only for MRC tasks, the BERT baseline was further improved as strong baseline by synthetic self training following \url{https://nlp.stanford.edu/seminar/details/jdevlin.pdf}.} The initial learning rate was set in \{8e-6, 1e-5, 2e-5, 3e-5\} with a warm-up rate of 0.1 and L2 weight decay of 0.01. The batch size was selected in \{16, 20, 32\}. The maximum number of epochs was set to 3 or 10 depending on tasks. The weight of $\alpha$ in the dual context aggregation was 0.5. All the texts were tokenized using wordpieces, and the maximum input length was set to 384 for both of SQuAD and RACE. The configuration for multi-head self-attention was the same as that for BERT. 

For the NLU task, the baseline is also the Whole Word Masking BERT. We used the pre-trained weights of BERT and followed the same fine-tuning procedure as BERT without any modification. The initial learning rate was set in the range \{2e-5, 3e-5\} with a warm-up rate of 0.1 and L2 weight decay of 0.01. The batch size was selected from \{16, 24, 32\}. The maximum number of epochs was set ranging from 2 to 5. Texts were tokenized using wordpieces, with a maximum length of 128.

For the NMT task, our baseline is Transformer \citep{NIPS2017_7181}. We used six layers for the encoder and the decoder. The number of dimensions of all input and output layers was set to 512. The inner feed-forward neural network layer was set to 2048. The heads of all multi-head modules were set to eight in both encoder and decoder layers. The byte pair encoding algorithm \cite{Sennrich2015Neural} was adopted, and the size of the vocabulary was set to 40,000. In each training batch, a set of sentence pairs contained approximately 4096$\times$4 source tokens and 4096$\times$4 target tokens. During training, the value of label smoothing was set to 0.1, and the attention dropout and residual dropout were \textit{p} = 0.1.  The Adam optimizer~\citep{DBLP:journals/corr/KingmaB14} was used to tune the parameters of the model.
The learning rate was varied under a warm-up strategy with 8,000 steps.
For evaluation, we validated the model with an interval of 1,000 batches on the dev set.
Following the training of 300,000 batches, the model with the highest BLEU score of the dev set was selected to evaluate the test sets.
During the decoding, the beam size was set to five. 

\begin{table}
\centering\setlength{\tabcolsep}{8.8pt}
\caption{\label{tab:squad2.0} Exact Match (EM) and F1 scores (\%) on SQuAD 2.0 dataset for single models. Our model is in boldface. $\dagger$ refers to unpublished works. Our final model is significantly better than the baseline BERT with p-value $<0.01$.}
	{
		\begin{tabular}{l c c c c }
			\hline
			\hline
			\multirow{2}{*}{\textbf{Model} }& \multicolumn{2}{c}{\textbf{Dev}} & \multicolumn{2}{c}{\textbf{Test}}\\
			& \textbf{EM} & \textbf{F1}&   \textbf{EM} & \textbf{F1}\\
			\hline
			\multicolumn{5}{c}{\emph{Regular Track}} \\
			Joint SAN &69.3 & 72.2 & 68.7 & 71.4\\	
			U-Net & 70.3  & 74.0  & 69.2 & 72.6 \\	
			RMR + ELMo + Verifier & 72.3  & 74.8 & 71.7 & 74.2 \\
			\hline
			\multicolumn{5}{c}{\emph{BERT Track}} \\
			Human& -  & - & 86.8&	89.5 \\
			\hdashline
			BERT \cite{devlin2018bert} & -  & - &  82.1 &  84.8 \\
			BERT + MMFT + ADA$\dagger$& -  & -  & 83.0 & 85.9 \\
			Insight-baseline-BERT$\dagger$& -  & - & 84.8 &	87.6 \\
			SemBERT \cite{zhang2019semantics} & -  & -  & 84.8 &	87.9   \\
			BERT + CLSTM + MTL + V$\dagger$& -  & -   & 84.9 &	88.2 \\
			BERT + NGM + SST$\dagger$& -  & - & 85.2 & 87.7\\
			BERT + DAE + AoA$\dagger$& -  & -   & 85.9 & 88.6 \\
			XLNet \cite{yang2019xlnet} & 87.9 & 90.6 & 87.9 & 90.7 \\
			ALBERT \cite{lan2019albert} & 87.4 & 90.2 & 88.1 & 90.9 \\
			\hline
			\multicolumn{5}{c}{\emph{Our implementation}} \\
			BERT Baseline & 84.1  & 86.8 & -  & - \\
			\textbf{SG-Net on BERT}  & 85.1 & 87.9 & - & -\\
			\quad\quad\textbf{+Verifier} &  85.6 & 88.3  & 85.2 & 87.9  \\
			\hdashline
			ALBERT Baseline & 87.0  & 90.1   & -  &  -  \\
			\textbf{SG-Net on ALBERT}  & 87.4 & 90.5 & - & -\\
			\quad\quad\textbf{+Verifier} & 88.0 & 90.8  & -  &  - \\
			\hline
			\hline
		\end{tabular}
	}
	
\end{table}

For span-based MRC, we conducted the significance test using McNemar's test \cite{mcnemar1947note} following \citeauthor{zhang2020retrospective}~(\citeyear{zhang2020retrospective}). For multi-choice MRC and NLI tasks, which are modeled as classification tasks, we performed statistical significance tests using paired t-test \cite{hsu2005paired}. For NMT, multi-bleu.perl was used to compute case-sensitive $4$-gram BLEU scores for NMT evaluations.\footnote{\url{https://github.com/moses-smt/mosesdecoder/tree/RELEASE-4.0/scripts/generic/multi-bleu.perl}} The signtest ~\citep{collins-koehn-kucerova:2005:ACL} is a standard statistical-significance test. All models were trained and evaluated on a single V100 GPU.

\begin{table}
\centering\setlength{\tabcolsep}{7.5pt}
		\caption{\label{tab:race} Accuracy (\%) on RACE test set for single models. Our model is significantly better than the baseline BERT with p-value $<0.01$.}
	{
		\begin{tabular}{l l l l}
			\hline
			\hline
			\textbf{Model} &   \textbf{RACE-M}  & \textbf{RACE-H} & \textbf{RACE}	\\
			\hline
			\multicolumn{4}{c}{\emph{Human Performance}} \\
			Turkers  & 85.1 & 69.4 & 73.3\\
			Ceiling  & 95.4 & 94.2& 94.5 \\
			\hline
			\multicolumn{4}{c}{\emph{Existing systems}} \\
			GPT \cite{radford2018improving} & 62.9 & 57.4 & 59.0 \\ 
			RSM \cite{sun2018improving}& 69.2 & 61.5 & 63.8  \\
			BERT \cite{ran2019option} & 75.6 & 64.7 & 67.9  \\
			OCN \cite{ran2019option} & 76.7 & 69.6 & 71.7  \\ 
			DCMN \cite{zhang2019dual}  & 77.6 & 70.1  & 72.3\\
			XLNet\cite{yang2019xlnet} & 85.5 & 80.2 & 81.8  \\
			ALBERT\cite{lan2019albert} & 89.0 & 85.5 & 86.5 \\ 
			\hline
			BERT Baseline & 78.4 & 70.4& 72.6 \\
			\textbf{SG-Net on BERT} & 78.8 & 72.2 & 74.2\\
			\hdashline
			ALBERT Baseline & 88.7 & 85.6 & 86.5 \\
			\textbf{SG-Net on ALBERT}  & 89.2 & 86.1 & 87.0 \\
			\hline
			\hline
		\end{tabular}
	}
\end{table}

\section{Experiments}
To focus on the evaluation of syntactic advance and keep simplicity, we only compare with single models instead of ensemble ones.

\subsection{Reading Comprehension}
Table \ref{tab:squad2.0} shows the result on SQuAD 2.0.\footnote{Besides published works, we also list competing systems on the SQuAD leaderboard at the time of submitting SG-Net (May 14, 2019). Our model is significantly better than the baseline BERT with p-value $<$ 0.01.} Various state of the art models from the official leaderboard are also listed for reference. We can see that the performance of BERT is very strong. However, our model is more powerful, boosting the BERT baseline essentially. It also outperforms all the published works and achieves the 2nd place on the leaderboard when submitting SG-NET. We also find that adding an extra answer verifier module could yield a better result. With the more powerful ALBERT, SG-Net still obtains improvements and also outperforms all the published works.

\begin{table}
	\centering
		{
		\caption{\label{tab:snli} Accuracy on the SNLI test set. Previous state-of-the-art models are marked by $\dagger$. Our model is significantly better than the baseline BERT with p-value $<0.05$.}}
		
	\begin{tabular}{p{6.5cm}p{0.5cm}}
		\hline
		\hline
		\textbf{Model} &  \textbf{Acc}   \\ 
		\hline
		\multicolumn{2}{c}{\emph{Existing systems}} \\
		GPT \cite{radford2018improving}  & 89.9 \\
		DRCN \cite{kim2018semantic} & 90.1\\
		MT-DNN$\dagger$ \cite{liu2019multi} & 91.6\\
		SemBERT$\dagger$ \cite{zhang2019semantics} & 91.6 \\
		\hline
		\multicolumn{2}{c}{\emph{Our implementation}} \\
		BERT Baseline & 91.5 \\
		\textbf{SG-Net} &  91.8\\
		\hdashline
		ALBERT Baseline & 92.6 \\
		\textbf{SG-Net on ALBERT} & 92.9 \\
		\hline
		\hline
	\end{tabular}
\end{table}

\begin{figure*}[!htb]
   \centering
   \subfigure{
   \begin{minipage}[b]{0.48\linewidth}
\setlength{\abovecaptionskip}{0pt}
\begin{center}
\pgfplotsset{height=5.6cm,width=8.2cm,compat=1.14,every axis/.append style={thick},every axis legend/.append style={ at={(0.95,0.95)}},every tick label/.append style={font=\scriptsize},legend columns=1 row=2} \begin{tikzpicture} \tikzset{every node}=[font=\small]

\begin{axis} [width=8.2cm,enlargelimits=0.13,legend pos=north west, xticklabels={2, 4, 6, 8, 10, 12, 14, 16, 18, 20, 22, 24, 26, 28}, axis y line*=left, axis x line*=left, xtick={0,1,2,3,4,5,6,7,8,9,10,11,12,13}, x tick label style={rotate=0},
  ylabel={Accuracy},
  ylabel style={align=left},xlabel={Question Length},font=\small]

\addplot+[sharp plot, mark=*,mark size=1.2pt,mark options={solid,mark color=red}, color=red] coordinates {(0, 82.05) (1, 84.58) (2, 86.97) (3, 87.78) (4, 86.95) (5, 87.61)  (6, 86.98) (7, 87.86) (8, 86.67) (9, 81.9)  (10, 82.69) (11, 82.35) (12, 100.0) (13, 90.0)};
\addlegendentry{\small SG-Net}

\addplot+ [sharp plot, mark=square*,mark size=1.2pt,mark options={mark color=cyan}, color=cyan] coordinates { (0, 61.54)  (1, 78.89) (2, 83.74) (3, 84.73) (4, 83.65) (5, 83.96) (6, 82.72) (7, 84.55) (8, 82.56)  (9, 82.86) (10, 80.77) (11, 64.71) (12, 81.82)  (13, 80.0)};
\addlegendentry{\small BERT}


\addplot+[densely dotted, mark=none, color=orange] coordinates {(0, 84.51)  (1, 84.69) (2, 85.06)  (3, 85.43) (4, 85.8)  (5, 86.17)  (6, 86.54) (7, 86.91) (8, 87.28) (9, 87.66) (10, 88.03) (11, 88.4) (12, 88.77)  (13, 89.14)};

\addplot+[densely dotted, mark=none, color=brown] coordinates {(0, 79.08)  (1, 79.14) (2, 79.25) (3, 79.36) (4, 79.47) (5, 79.58) (6, 79.69) (7, 79.8) (8, 79.91) (9, 80.02) (10, 80.13) (11, 80.24)  (12, 80.35) (13, 80.46)};

\end{axis}
\end{tikzpicture}

\end{center}
   \centerline{(a) Split by equal range of question length}
   \end{minipage}
   }
   \subfigure{
   \begin{minipage}[b]{0.48\linewidth}
\setlength{\abovecaptionskip}{0pt}
\begin{center}

\pgfplotsset{height=5.6cm,width=8.2cm,compat=1.14,every axis/.append style={thick},every tick label/.append style={font=\scriptsize},every axis legend/.append style={ at={(0.95,0.95)}},legend columns=1 row=2} 

\begin{tikzpicture} \tikzset{every node}=[font=\small] 
 \begin{axis} [width=8.2cm,enlargelimits=0.13,legend pos=north west, xticklabels={4, 5, 6,  , 7,  , , 8,  , 9,  ,  , 10,  , 11, 12,  , 13, 15, 18, 28}, axis y line*=left,
axis x line*=left, xtick={0,1,2,3,4,5,6,7,8,9,10,11,12,13,14,15,16,17,18,19,20}, 
x tick label style={rotate=0},
  ylabel={Accuracy},
  ylabel style={align=left},xlabel={Question Length},font=\small]

\addplot+[sharp plot, mark=*,mark size=1.2pt,mark options={solid,mark color=red}, color=red] coordinates { (0, 84.82)  (1, 84.32)  (2, 87.02)  (3, 87.86)  (4, 88.53)  (5, 85.83)  (6, 88.36)  (7, 87.35)  (8, 88.03)  (9, 89.04)  (10, 86.51)  (11, 86.68)  (12, 86.51)  (13, 86.51)  (14, 87.69)  (15, 88.2)  (16, 88.7)  (17, 86.85)  (18, 86.85)  (19, 85.83)  (20, 92.31)};
\addlegendentry{\small SG-Net}

\addplot+ [sharp plot, mark=square*,mark size=1.2pt,mark options={mark color=cyan}, color=cyan] coordinates {(0, 76.73)  (1, 81.28)  (2, 83.14)  (3, 85.33)  (4, 85.83)  (5, 83.47)  (6, 85.67)  (7, 83.64)  (8, 84.99)  (9, 86.0)  (10, 82.12)  (11, 84.65)  (12, 82.46)  (13, 83.81)  (14, 82.63)  (15, 85.16)  (16, 85.5)  (17, 82.97) (18, 82.46) (19, 82.97)  (20, 76.92)};
\addlegendentry{\small BERT}

\addplot+[densely dotted, mark=none, color=orange] coordinates {(0, 83.44)  (1, 83.42)  (2, 83.4)  (3, 83.38)  (4, 83.35)  (5, 83.33)  (6, 83.31)  (7, 83.29)  (8, 83.27)  (9, 83.25) (10, 83.23)  (11, 83.2)  (12, 83.18)  (13, 83.16)  (14, 83.14)  (15, 83.12)  (16, 83.1)  (17, 83.08)  (18, 83.06)  (19, 83.03)  (20, 83.01)};

\addplot+[densely dotted, mark=none, color=brown] coordinates {(0, 86.26)  (1, 86.36)  (2, 86.47)  (3, 86.58)  (4, 86.68)  (5, 86.79)  (6, 86.9)  (7, 87.0)  (8, 87.11)  (9, 87.22)  (10, 87.32)  (11, 87.43)  (12, 87.54)  (13, 87.64)  (14, 87.75)  (15, 87.86)  (16, 87.96)  (17, 88.07)  (18, 88.18)  (19, 88.28)  (20, 88.39)};
\end{axis}

\end{tikzpicture}
\end{center}
   \centerline{(b) Split by equal amount of questions}
   \end{minipage}
   }
\caption{\label{fig:length} Accuracy for different question length. Each data point means the accuracy for the questions in the same length range (a) or of the same number (b) and the horizontal axis in (b) shows that most of questions are of length 7-8 and 9-10.}
\end{figure*}

\begin{figure*}[!htb]
   \centering
   \subfigure{
   \begin{minipage}[b]{0.48\linewidth}
\setlength{\abovecaptionskip}{0pt}
\begin{center}
\pgfplotsset{height=5.6cm,width=8.2cm,compat=1.14,every axis/.append style={thick},every tick label/.append style={font=\scriptsize},every axis legend/.append style={ at={(0.95,0.95)}},legend columns=1 row=2} \begin{tikzpicture} \tikzset{every node}=[font=\small]

\begin{axis} [width=8.2cm,enlargelimits=0.13,legend pos=north west, xticklabels={10,20,30,40,50,60,70,80,80+}, axis y line*=left, axis x line*=left, xtick={0,1,2,3,4,5,6,7,8}, x tick label style={rotate=0},
  ylabel={BLEU},
  ylabel style={align=left},xlabel={Sentence Length},font=\small]

\addplot+[sharp plot, mark=*,mark size=1.2pt,mark options={solid,mark color=red}, color=red] coordinates {(0, 21.04)  (1, 22.88)  (2, 22.24)  (3, 23.01)  (4, 21.88)  (5, 22.44)  (6, 27.59)  (7, 14.37)  (8, 27.28)};
\addlegendentry{\small SG-Net}

\addplot+ [sharp plot, mark=square*,mark size=1.2pt,mark options={mark color=cyan}, color=cyan] coordinates {(0, 20.53)  (1, 22.16)  (2, 21.69)  (3, 22.93)  (4, 21.34)  (5, 21.70)  (6, 25.37)  (7, 13.16)  (8, 23.87)};
\addlegendentry{\small Transformer}


\addplot+[densely dotted, mark=none, color=orange] coordinates {(0, 21.89)(1, 22.05)(2, 22.21) (3, 22.37) (4, 22.53)(5, 22.68)(6, 22.84) (7, 23.0) (8, 23.16)};

\addplot+[densely dotted, mark=none, color=brown] coordinates {(0, 21.92)(1, 21.79)(2, 21.67)(3, 21.54)(4, 21.42)(5, 21.29)(6, 21.17)(7, 21.04)(8, 20.92)};

\end{axis}
\end{tikzpicture}

\end{center}
   \centerline{(a) Word-level NMT}
   \end{minipage}
   }
   \subfigure{
   \begin{minipage}[b]{0.48\linewidth}
\setlength{\abovecaptionskip}{0pt}
\begin{center}

\pgfplotsset{height=5.6cm,width=8.2cm,compat=1.14,every axis/.append style={thick},every tick label/.append style={font=\scriptsize},every axis legend/.append style={ at={(0.95,0.95)}},legend columns=1 row=2} 

\begin{tikzpicture} \tikzset{every node}=[font=\small] 
 \begin{axis} [width=8.2cm,enlargelimits=0.13,legend pos=north west, xticklabels={10,20,30,40,50,60,70,80,80+}, axis y line*=left,
axis x line*=left, xtick={0,1,2,3,4,5,6,7,8}, 
x tick label style={rotate=0},
  ylabel={BLEU},
  ylabel style={align=left},xlabel={Sentence Length},font=\small]

\addplot+[sharp plot, mark=*,mark size=1.2pt,mark options={solid,mark color=red}, color=red] coordinates {(0, 25.91)  (1, 26.67)  (2, 26.12)  (3, 27.95)  (4, 27.26)  (5, 27.69)  (6, 30.75)  (7, 24.67)  (8, 40.11)};
\addlegendentry{\small SG-Net}

\addplot+ [sharp plot, mark=square*,mark size=1.2pt,mark options={mark color=cyan}, color=cyan] coordinates {(0, 25.77)  (1, 26.51)  (2, 25.61)  (3, 27.67)  (4, 26.73 )  (5, 26.42)  (6, 29.00)  (7, 22.40)  (8, 35.98)};
\addlegendentry{\small Transformer}

\addplot+[densely dotted, mark=none, color=orange] coordinates {(0, 24.58)(1, 25.58)(2, 26.58)(3, 27.57)(4, 28.57)(5, 29.57)(6, 30.56)(7, 31.56)(8, 32.56)};

\addplot+[densely dotted, mark=none, color=brown] coordinates {(0, 25.07)(1, 25.64)(2, 26.21)(3, 26.78)(4, 27.34)(5, 27.91)(6, 28.48)(7, 29.05)(8, 29.61)};
\end{axis}

\end{tikzpicture}
\end{center}
   \centerline{(b) Subword-level NMT }
   \end{minipage}
   }
\caption{\label{fig:nmt_length} Translation  qualities  of  different  sentence lengths. Each data point means the BLEU score in (a) word-level or (b) subword-level.}
\end{figure*}

For RACE, we compare our model with the following latest baselines: Dual Co-Matching Network (DCMN) \cite{zhang2019dual}, Option Comparison Network (OCN) \cite{ran2019option}, Reading Strategies Model (RSM) \cite{sun2018improving}, and Generative Pre-Training (GPT) \cite{radford2018improving}. Table \ref{tab:race} shows the result.\footnote{Our concatenation order of $P$ and $Q$ is slightly different from the original BERT. Therefore, the result of our BERT baseline is higher than the public one on the leaderboard, thus our improved BERT implementation is used as the stronger baseline for our evaluation.} Turkers is the performance of Amazon Turkers on a random subset of the RACE test set. Ceiling is the percentage of unambiguous questions in the test set. From the comparison, we can observe that our model outperforms all baselines, which verifies the effectiveness of our proposed syntax enhancement.

\subsection{Natural Language Inference}
Table \ref{tab:snli} shows SG-Net also boosts the baseline BERT, and achieves a new state-of-the-art on SNLI benchmark. The result shows that using our proposed method with induced syntactic structure information can also benefit natural language inference.

\begin{table}
\centering
{
	{
		\caption{BLEU scores on EN-DE dataset for the NMT tasks. ``+/++" after the score indicates that the proposed method was significantly better than the baseline at significance level $p<0.05/0.01$.}
		\label{tbl:nmt}} 
		
	\begin{tabular}{p{3.5cm}p{1cm}p{1cm}}
		\hline
		\hline
		\textbf{Model} &  \textbf{Word} & \textbf{Subword}  \\ 
		\hline
		\multicolumn{3}{c}{\emph{Existing systems}} \\
		Transformer \cite{NIPS2017_7181} & -  & 27.3 \\
		SDAtt \cite{chen2018syntax} & 20.36 & - \\
		\hline
		\multicolumn{3}{c}{\emph{Our implementation}} \\
		Transformer & 22.06 & 27.31   \\
		\textbf{SG-Net} & \textbf{23.02}++ & \textbf{27.68}+ \\
		\hline
		\hline
	\end{tabular}
	}

\end{table}

\subsection{Neural Machine Translation}
Table~\ref{tbl:nmt} shows the translation results. Our method significantly outperformed the baseline Transformer. The results showed that our method was not only useful for the NLU task but also more advanced translation tasks.

\begin{figure*}
	\centering
	\includegraphics[width=0.95\textwidth]{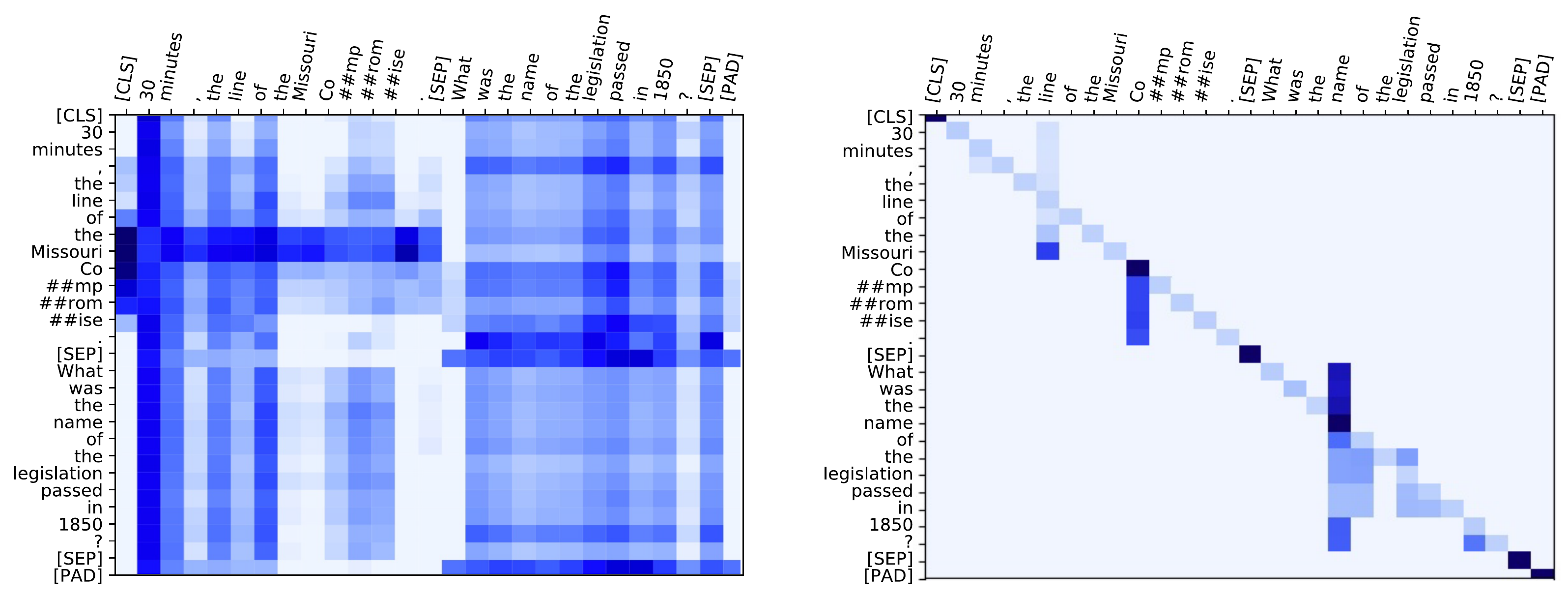}
	\\\begin{flushleft}
		\scriptsize{\emph{
				\textbf{Passage (extract)}:...30 minutes, the line of the Missouri Compromise... \textbf{Question}:What was the name of the legislation passed in 1850?  \textbf{Answer}:the Missouri Compromise} }
	\end{flushleft}
	\caption{\label{fig:vis-att} Visualization of the vanilla BERT attention (left) and syntax-guided self-attention (right). Weights of attention are selected from first head of the last attention layer. For the syntax-guided self-attention, the columns with weights represent the SDOI for each word in the row. For example, the SDOI of \textit{passed} contains \textit{\{name, of, legislation, passed\}}. Weights are normalized by SoftMax for each row. }
\end{figure*}

\section{Discussions}

\subsection{Effect of Answering Long Questions}
\label{sec:long}
Inspired by previous works that introducing syntactic structures might be beneficial for the modeling of long sentences \cite{chen2018syntax}, we are interested in whether the advance still holds when it comes to Transformer architectures. We grouped  questions/sentences  of  similar lengths for both of the passage-level MRC and sentence-level NMT tasks, to investigate the model performance. For example, sentence length ``10" indicates  that  the  length  of questions  is between 8 and 10.

In detail, we sort the questions from the SQuAD 2.0 dev set according to the length and then group them into 20 subsets, split by equal range of question length and equal amount of questions. Then we calculate the exact match accuracy of the baseline and SG-Net per group, as shown in Figure \ref{fig:length}. 

Since the number of questions varies in length, the gap could be biased by the distribution. We depict the two aspects to show the discovery comprehensively and we would like to clarify that the trends (dashed) in Figure \ref{fig:length} matter most, giving the intuition that the proposed one might not be as sensitive to long sentences as the baseline. We observe that the performance of the baseline drops heavily when encountered with long questions, especially for those longer than 20 words while our proposed SG-Net works robustly, even showing a positive correlation between accuracy and length. This shows that with syntax-enhanced representation, our model is better at dealing with lengthy questions compared with baseline.

For sentence-level NMT, we illustrate the comparison for both the word-level and subword-level NMT settings. According to the illustrations in Figure \ref{fig:nmt_length}, we see similar trends with the passage-level MRC. We observe that our method can still boost the baseline substantially though the Transformer baseline has good ability of long-dependency modeling. We also find that SG-Net shows relatively greater advance for the passage-level task than that of sentence-level, in which there are always hundreds of words in an input sentence for passage-level MRC, which would require the stronger capacity of long-dependency, as well as structure modeling.

\subsection{Visualization}
To have an insight that how syntax-guided attention works, we draw attention distributions of the vanilla attention of the last layer of BERT and our proposed syntax-guided self-attention,\footnote[11]{Since special symbols such as [PAD] and [CLS] are not considered in the dependency parsing tree, we confine the SDOI of these tokens to themselves. So these special tokens will have a value of 1 as weights over themselves in syntax-guided self-attention, and we will mask these weights in the following aggregation layer.} as shown in Figure \ref{fig:vis-att}. With the guidance of syntax, the keywords \emph{name}, \emph{legislation} and \emph{1850} in the question are highlighted, and \emph{(the)} \emph{Missouri}, and \emph{Compromise} in the passage are also paid great attention, which is exactly the right answer. The visualization verifies that benefiting from the syntax-guided attention layer; our model is effective at selecting the vital parts, guiding the downstream layer to collect more relevant pieces to make predictions.

\begin{table}
	\centering
		\caption{\label{tab:ablation} Ablation study on potential components and aggregation methods on SQuAD 2.0 dev set.}
	\begin{tabular}{l c c}
		\hline
		
		\hline
		\textbf{Model} & \textbf{EM} & \textbf{F1}\\
		\hline
		baseline & 84.1 & 86.8 \\
		+ Vanilla attention only & 84.2 &  86.9 \\
		+ Syntax-guided attention only & 84.4 & 87.2  \\
		+ Dual contextual attention & \textbf{85.1} & \textbf{87.9} \\
		\hdashline
		Concatenation &   84.5  & 87.6 \\	
		Bi-attention & 84.9 & 87.8\\
		\hline
	\end{tabular}
\end{table}

\subsection{Dual Context Mechanism Evaluation}
In SG-Net, we integrate the representations from syntax-guided attention layer and the vanilla self-attention layer in dual context layer. To unveil the contribution of each potential component, we conduct comparisons on the baseline with:

\begin{enumerate}
    \item \emph{Vanilla attention only} that adds an extra vanilla BERT attention layer after the BERT output.
    \item \emph{Syntax-guided attention only} that adds an extra syntax-guided layer after the BERT output.
    \item \emph{Dual contextual attention} that is finally adopted in SG-Net as described in Section \ref{SG_ATT}.
\end{enumerate}

\begin{table*}

\centering
	{
		\caption{Eight probing tasks \cite{conneau2018you} to study what kind of properties are captured by the encoders.}
		\label{tbl:prob_type}} 
	{
	{
\begin{tabular}{p{1.5cm}|p{1.5cm}|p{13cm}}
		\hline
		\hline
		\multicolumn{2}{c|}{Probing Tasks} &  Content \\ 
		\hline
        \multirow{3}{*}{Syntactic} & TrDep & Checking whether an encoder infers the hierarchical structure of sentence \\
        \cline{2-3}
        & \multirow{2}{*}{ToCo} & Sentences should be classified in terms of the sequence of top constituents immediately below the sentence node \\
        \cline{2-3}
        & BShif & Testing whether two consecutive tokens within the sentence have been inverted \\
        \hline
        \multirow{5}{*}{Semantic} & Tense & Asking for the tense of the main clause verb \\
        \cline{2-3}
        & SubN & Focusing on the number of the main clause’s subject\\
        \cline{2-3}
        & ObjN & Testing for the number of the direct object of the main clause \\
        \cline{2-3}
        & \multirow{2}{*}{SoMo} & Some sentences are modified by replacing a random noun or verb with another one and the classifier should tell whether a sentence has been modified  \\
        \cline{2-3}
        & CoIn & Containing sentences made of two coordinate clauses \\
		\hline
		\hline
	\end{tabular}
	}
	}

\end{table*}

\begin{table*}
\setlength{\tabcolsep}{8pt}
\centering
	{
		\caption{Accuracy on eight probing tasks of evaluating linguistics embedded in the encoder outputs.  ``++/+" after the accuracy score indicate that the score is statistically significant at level $p < 0.01/0.05$.}
		\label{tbl:prob_result}} 
	{
	\begin{tabular}{l|lll|lllll}
		\hline
		\hline
		\multirow{2}{*}{Model} & \multicolumn{3}{c|}{Syntactic} & \multicolumn{5}{c}{Semantic} \\
		& TrDep & ToCo & BShif & Tense & SubN & ObjN & SoMo & CoIn \\
		\hline
        Baseline & 28.34  & 58.33 & 76.34& 80.66 & 77.28 & 76.32 & 64.42 & 67.51 \\
        SG-Net & 30.02++  &  60.42++ & 76.63  & 80.52 & 77.72+ & 76.83+ & 64.38 & 67.23\\
		\hline
		\hline
	\end{tabular}
	}
\end{table*}

\begin{table*}
	\centering
		\caption{\label{tab:ans}The comparison of answers from baseline and our model. In these examples, answers from SG-Net are the same as the ground truth.}
	{
		\begin{tabular}{l c c}
			\hline
			
			\hline
			Question & Baseline  & SG-Net\\
			\hline
			When did Herve serve as a Byzantine general? & 105 & 1050s\\	
			What is Sky+ HD material broadcast using? &MP  &MPEG-4\\ 
			A problem set that that is hard for the expression NP can also be stated how? & the set of NP-hard problems & NP-hard\\
			\hline
			
			\hline
		\end{tabular}
	}

\end{table*}

Table \ref{tab:ablation} shows the results. We observe that dual contextual attention yields the best performance. Adding extra vanilla attention gives no advance, indicating that introducing more parameters would not promote the strong baseline. It is reasonable that syntax-guided attention only is also trivial since it only considers the syntax related parts when calculating the attention, which is complementary to traditional attention mechanism with noisy but more diverse information and finally motivates the design of dual contextual layer. 

Actually, there are other operations for merging representations in dual context layer besides the weighted dual aggregation, such as \emph{concatenation} and \emph{Bi-attention} \cite{Seo2016Bidirectional}, which are also involved in our comparison, and our experiments show that using dual contextual attention produces the best result.

\subsection{Linguistic Analysis}

We are interested in what kind of knowledge learned in the universal representations. In this section, we selected eight widely-used language probing tasks \cite{conneau2018you} (see Table \ref{tbl:prob_type}) to study what syntactic and semantic properties are captured by the encoders. Specifically, we use the BERT and ALBERT models trained on the SNLI task to generate the sentence representation (embedding) of input to evaluate what linguistic properties are encoded in it. The results are shown in Table \ref{tbl:prob_result}. Regarding the syntactic properties, our model achieves statistically significant improvement on the \textit{TrDep} and \textit{ToCo} tasks, which demonstrates that the syntax-guided model could help identify and represent the sentence structure and constituents, which discloses that SG-Net might learn better syntax-aware information with the syntax-constrained SAN design.

\subsection{Model Prediction}\label{span_out}
To have an intuitive observation of the predictions of SG-Net, we show a list of prediction examples on SQuAD 2.0 from baseline BERT and SG-Net in Table \ref{tab:ans}. The comparison indicates that our model could extract more syntactically accurate answer, yielding more exact match answers while those from the baseline BERT model are often semantically incomplete. This shows that utilizing explicit syntax is potential to guide the model to produce meaningful predictions. Intuitively, the advance would attribute to better awareness of syntactic dependency of words, which guides the model to learn the syntax relationships explicitly. 

\subsection{Influence of Parsing Performance}\label{sec:parse_acc}

Since our SDOI mask is derived from the syntactic parsing, the parsing performance would influence the overall model performance. To investigate the influence of the accuracy of the parser, we first use a relatively weaker parser with 95.62\% UAS and 94.10\% LAS accuracy, using other suboptimal hyper-parameters. We use the NLI task on SNLI for benchmark, the accuracy is 91.72\% compared with the best result of 91.81\% reported in Table \ref{tab:snli}. To evaluate in more extreme cases, we then degrade our parser by randomly turning specific proportion [0, 20\%, 40\%, 60\%, 80\%, 100\%] of labels into random error ones as cascading errors. The SNLI accuracies are respectively [91.81\%, 91.68\%, 91.56\%, 91.42\%, 91.41\%, 91.38\%]. This stable performance can be attributed to dual attention design, i.e., the concatenation operation of BERT hidden states and pruned syntax-guided layer representation, in which the downgraded parsing representation (even noisy)  would not affect the former one intensely. This result indicates that the LM can not only benefit from high-accuracy parser but also keep robust against noisy labels.

\subsection{Model Efficiency}
There are two aspects that contribute to the overall model computation: 1) preprocessing: the syntactic parsing on our task datasets; and 2) model training: training SG-Net for our tasks. Our calculation analysis is based on one 32G V100 GPU. For the processing, the annotation speed is rapid, which is around 10.3 batches (batch size=256 sentences, max sequence length=256) per second. For model training, the number of extra trainable parameters in our overall model is very small (only +12M, compared with 335M of BERT, for example) as we only add one extra attention layer. The dual aggregation is the tensor calculation on the existing outputs of SG layer and BERT, without any additional parameter. The training computation speeds of SG-Net and the baseline are basically the same (1.18 vs. 1.17 batches per second, batch size=32, max sequence length=128).

\section{Discussion of Pros and Cons}
There are a variety of methods proposed to introduce tree or graph-based syntax into RNN-based NLP models. The main idea of the traditional solutions is transforming the data form by either 1) applying a tree-based network \cite{ma2018forest}, or 2) linearizing syntax \cite{li2017modeling,eriguchi2016tree}, which converts the relationship of the nodes in dependency tree into discrete labels and integrates the label-formed features with word embeddings. The difference lies with the efficiency and the processing part in the neural models. Among the two methods, the advantage of the former is that it can preserve the original structured characteristics of the data, i.e., the attributes of words and relationships between different words, and the tree network can fully display the tree structure in the network. In contrast, the latter's benefits are simplicity and high efficiency, where the syntactic features can be easily integrated with existing word embeddings. In terms of the processing part, the former will perform in the sentence-level encoding phase, while the latter focuses on the fundamental embedding layer before contextual encoding. 

For the recent Transformer architecture, the self-attention mechanism ensures that the model can fully obtain the dependency between each token of the input sequence, which goes beyond the sequential processing of each successive token. Such a relationship is richer and more diverse than RNN models. However, Transformer could not score the relationship between tokens explicitly. Although the relationship might be obtained to some extent after training, the adjustment could be chaotic, redundant and unexplained. Syntactic information, which can reflect the relationship between tokens, is obviously a natural solution as a kind of guidance. Compared with the traditional methods, SG-Net has the following advantages:

1. High precision. The intermediate product of the attention matrix generated by the self-attention can be regarded as a set of relations among each couple of tokens in size of $n \times n$ where $n$ denotes the sequence length, that is, a complete graph as well. SG-Net can well save the tree or graph form of features to the greatest extent, and it is intuitive and interpretable. 

2. High efficiency. Compared with the traditional processing of tree-based networks, SG-Net keeps the parallelism of Transformer with minor revisions in the architecture. Besides, the data processing is finished before the model training, instead of putting the data processing in the model training as in traditional methods.

3. Light-weight. SG-Net makes full use of the structure of Transformer, and the change has little damage to Transformer and can be fully integrated into Transformer. 

However, there still exist some disadvantages in theory. First, SG-Net can only be used in Transformer architectures that rely on self-attention mechanisms. Second, the type of relationship is considered. Although it can reflect the relationship between tokens, it can not reflect what kind of relationship it is. Third, compared with the traditional method, SG-Net requires high-precision syntax, and the wrong syntax will cause great damage. The reason is that this method directly modifies the strength of the relationship between tokens, which can suffer from error propagation. However, as discussed in Section \ref{sec:parse_acc}, our dual attention design can well alleviate such a situation when using a weak parser and ensure the model focus more on the original output.

\section{Conclusion}
This paper presents a novel syntax-guided framework for enhancing strong Transformer-based encoders. We explore to adopt syntax to guide the text modeling by incorporating syntactic constraints into attention mechanism for better linguistically motivated word representations. Thus, we adopt a dual contextual architecture called syntax-guided network (SG-Net) which fuses both the original SAN representations and syntax-guided SAN representations. Taking pre-trained BERT as our Transformer encoder implementation, experiments on three typical advanced natural language understanding tasks, including machine reading comprehension, natural language inference, and neural machine translation show that our model can yield significant improvements in all the challenging tasks. This work empirically discloses the effectiveness of syntactic structural information for text modeling. The proposed attention mechanism also verifies the practicability of using linguistic information to guide attention learning and can be easily adapted with other tree-structured annotations.

For the application, our method is lightweight and easy to cooperate with other neural models. Without the need for architecture modifications in existing models, we can easily apply syntax-guided constraints in an encoder by adding an extra layer fed by the output representation of the encoder and the SDOI mask, which makes it easy enough to be applicable to other systems. The design is especially important for recent dominant pre-trained Transformer-based language representation models, such as BERT \cite{devlin2018bert}, XLNet \cite{yang2019xlnet}, RoBERTa \cite{liu2019roberta}, ALBERT \cite{lan2019albert}, etc. As a plugin, we do not need to train the models from scratch but directly fine-tune for downstream tasks.

Incorporating human expertise and knowledge to machine is a key inspiration of AI researches, which stimulates lots of studies that consider involving linguistic information to neural models in NLP scenarios. In this work, we discovered an effective masking strategy to incorporate extra structured knowledge into the state-of-the-art Transformer architectures. Besides syntactic information, this method is also compatible with a variety of structured knowledge as explicit sentence-level constraints to improve the representation ability of Transformer. For example, it is potential to model the concept relationships by introducing extra structured knowledge graphs, including ConceptNet \cite{speer2017conceptnet}, DBpedia \cite{auer2007dbpedia}. We hope this work can facilitate related research and shed lights on future studies of knowledge aggregation in the community.

\ifCLASSOPTIONcaptionsoff
  \newpage
\fi

\bibliographystyle{IEEEtranN}
\bibliography{sg-net}

\vspace{-12mm}
	\begin{IEEEbiography}[{\includegraphics[width=1in,height=1.25in,clip,keepaspectratio]{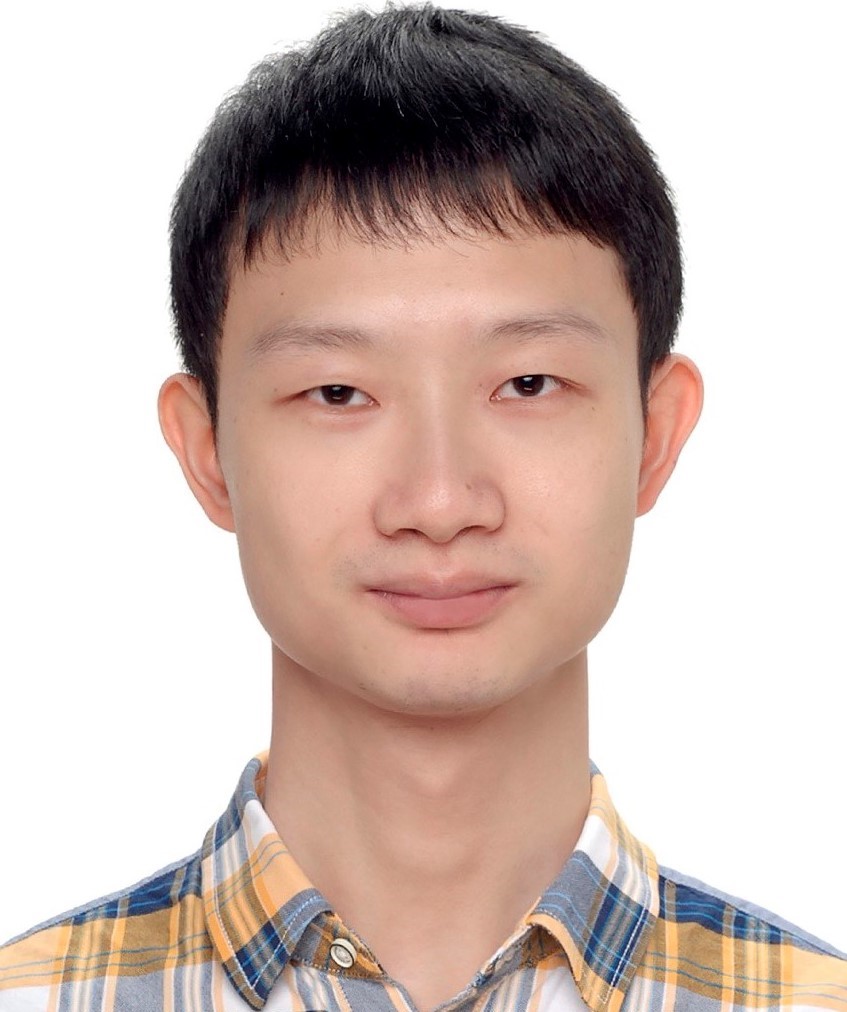}}]{Zhuosheng Zhang}
  		received his Bachelor's degree in internet of things from Wuhan University in 2016, his M.S. degree in computer science from Shanghai Jiao Tong University in 2020. He is working towards the Ph.D. degree in computer science with the Center for Brain-like Computing and Machine Intelligence of Shanghai Jiao Tong University. He was an internship research fellow at NICT from 2019-2020. His research interests include natural language processing, machine reading comprehension, dialogue systems, and machine translation. 
	\end{IEEEbiography}
	
\vspace{-12mm}
	\begin{IEEEbiography}[{\includegraphics[width=1in,height=1.25in,clip,keepaspectratio]{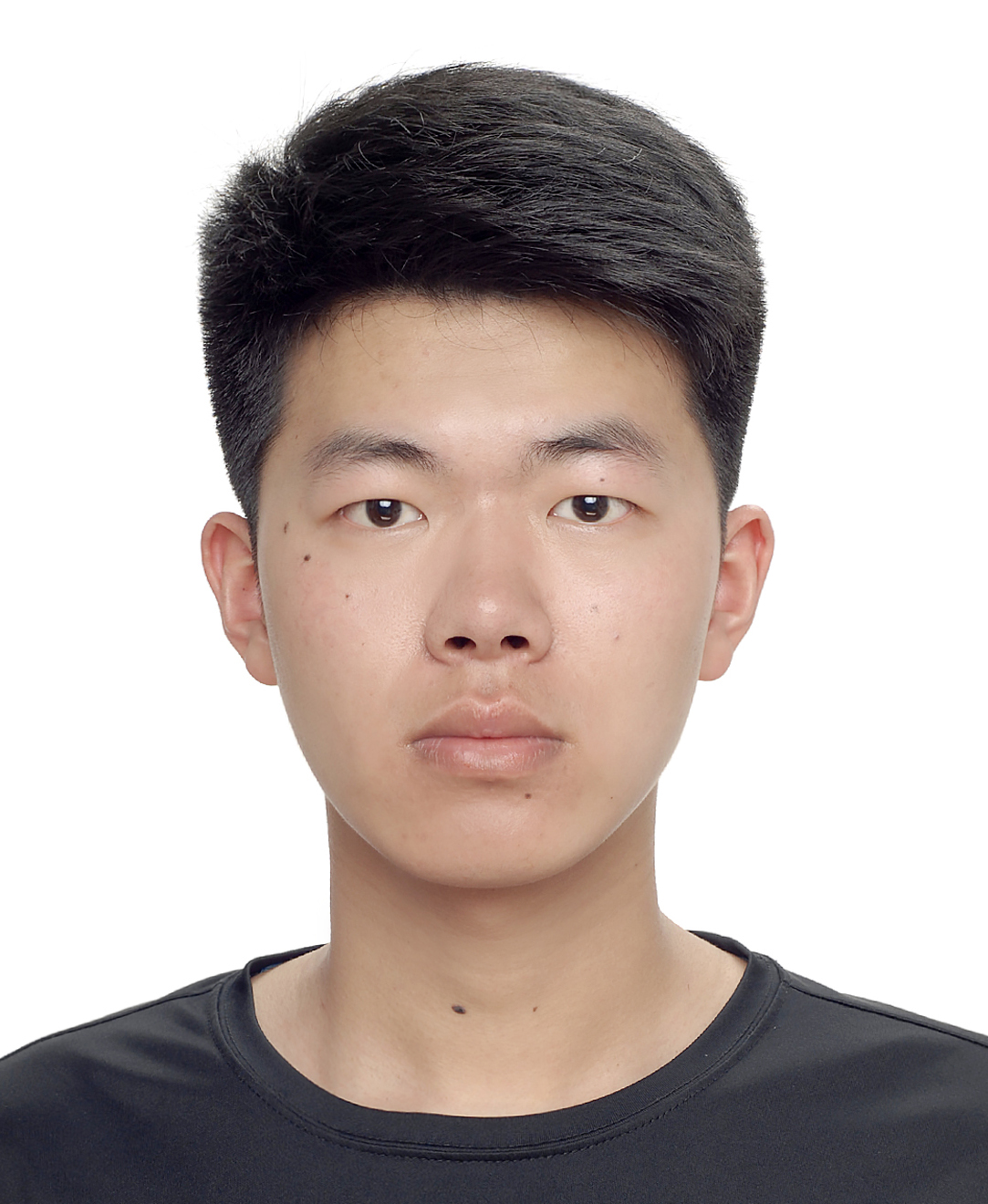}}]{Yuwei Wu}
  	received his Bachelor's degree in Computer Science from Shanghai Jiao Tong University in 2020, where he was a member of ACM Class, part of Zhiyuan College in SJTU. He is working towards the Ph.D. degree in Department of Computer Science, Shanghai Jiao Tong University. During his graduate study, he took research internships at Georgia Tech in 2019. His research interests lie in natural language processing, machine reading comprehension, dialogue systems, and multimodal.
    \end{IEEEbiography}
	
\vspace{-12mm}
	\begin{IEEEbiography}[{\includegraphics[width=1in,height=1.25in,clip,keepaspectratio]{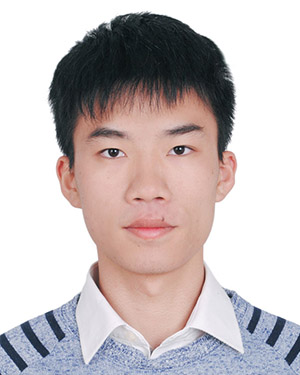}}]{Junru Zhou}
  	 received the B.S. degree from South China University of Technology, in 2018.
Since then, he has been a master student in Department of Computer Science and Engineering, Shanghai Jiao Tong University. His research focuses on natural language processing, especially in syntactic and semantic parsing, pre-training language model.
    \end{IEEEbiography}
    
\vspace{-12mm}
	\begin{IEEEbiography}[{\includegraphics[width=1in,height=1.25in,clip,keepaspectratio]{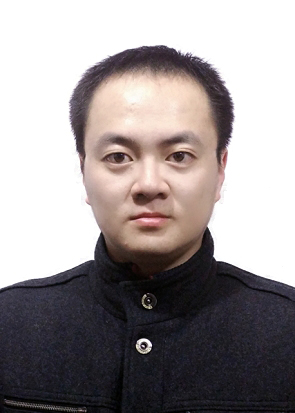}}]{Sufeng Duan}
   received his Bachelor degree in Spatial Information and Digital Technology from Wuhan University and Master degree in Computer Science and Technology from Shanghai Jiao Tong University in 2014 and 2017. Since 2018, he has been a Ph.D. student with the Center for Brain-like Computing and Machine Intelligence of Shanghai Jiao Tong University, Shanghai, China. His research focuses on natural language processing, especially Chinese word segmentation and machine translation.
    \end{IEEEbiography}
    
\vspace{-12mm}
	\begin{IEEEbiography}[{\includegraphics[width=1in,height=1.25in,clip,keepaspectratio]{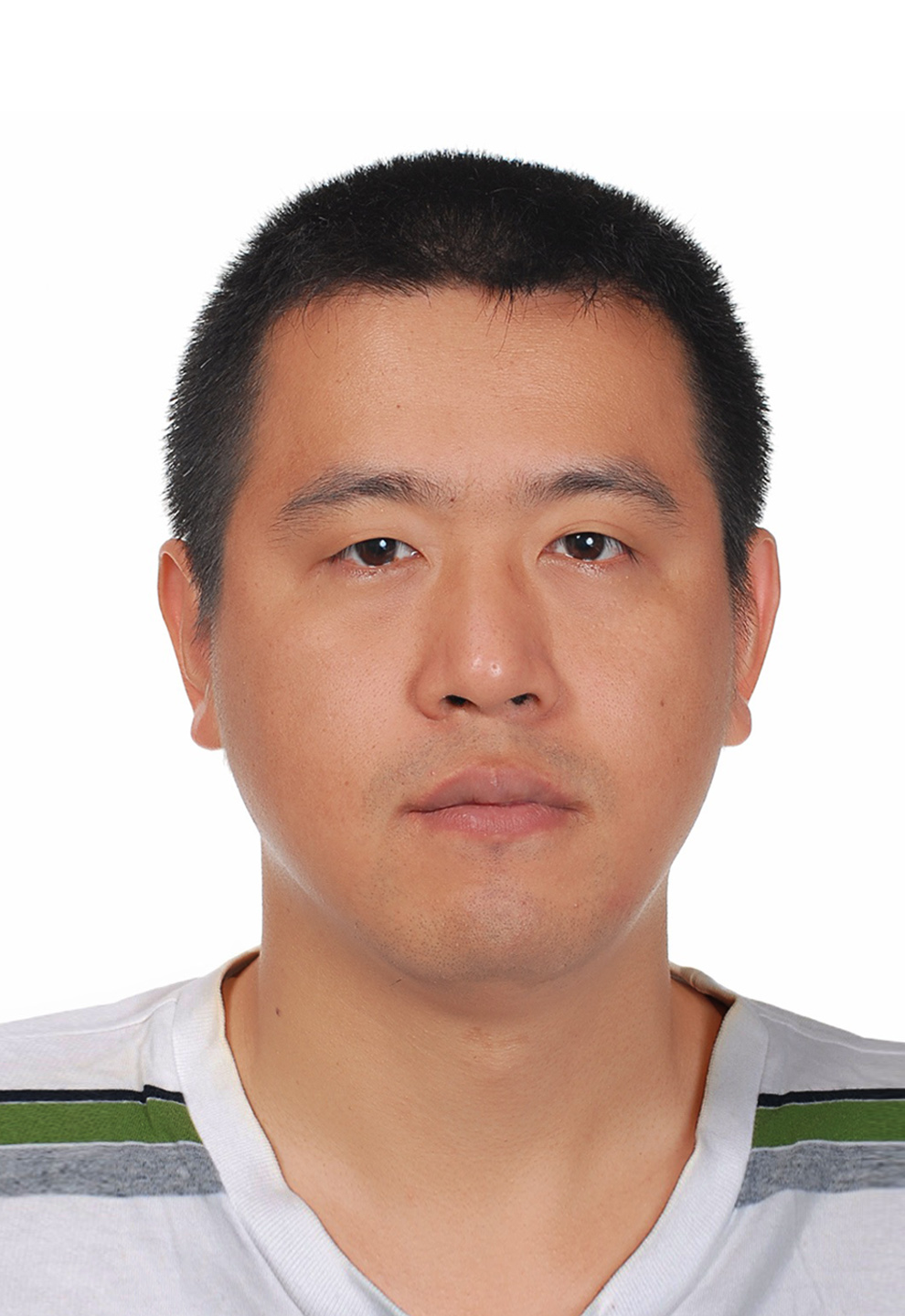}}]{Hai Zhao}
		received the BEng degree in sensor and instrument engineering, and the MPhil degree in control theory and engineering from Yanshan University in 1999 and 2000, respectively,
		and the PhD degree in computer science from Shanghai Jiao Tong University, China in 2005. 
		He is currently a full professor at department of computer science and engineering,  Shanghai Jiao Tong University after he joined the university in 2009. 
		He was a research fellow at the City University of Hong Kong from 2006 to 2009, a visiting scholar in Microsoft Research Asia in 2011, a visiting expert in NICT, Japan in 2012.
		He is an ACM professional member, and served as area co-chair in ACL 2017 on Tagging, Chunking, Syntax and Parsing, (senior) area chairs in ACL 2018, 2019 on Phonology, Morphology and Word Segmentation.
		His research interests include natural language processing and related machine learning, data mining and artificial intelligence.
	\end{IEEEbiography}
	
\vspace{-12mm}
\begin{IEEEbiography}[{\includegraphics[width=1in,height=1.25in,clip,keepaspectratio]{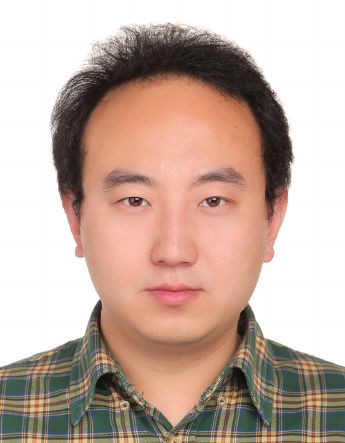}}]{Rui Wang}
	is an associate professor at Shanghai Jiao Tong University since 2021. Before that, he was a researcher (tenured in 2020) at Japan National Institute of Information and Communications Technology (NICT) from 2016 to 2020. He received his B.S. degree from Harbin Institute of Technology in 2009, his M.S. degree from the Chinese Academy of Sciences in 2012, and his Ph.D. degree from Shanghai Jiao Tong University in 2016, all of which are in computer science. He was a joint Ph.D. at Centre Nationnal de la Recherche Scientifique, France in 2014. His research interests are machine translation and natural language processing.
\end{IEEEbiography}

\end{document}